\documentclass[preprint]{article}

\usepackage{neurips_2023}

\usepackage[utf8]{inputenc} 
\usepackage[T1]{fontenc}    
\usepackage{hyperref}       
\usepackage{url}            
\usepackage{booktabs}       
\usepackage{amsfonts}       
\usepackage{nicefrac}       
\usepackage{microtype}      
\usepackage{xcolor}         
\usepackage{listings}
\usepackage{cite}
\usepackage{graphicx}
\usepackage{hyperref}
\usepackage{amsmath}

\hypersetup{
    pdftitle={Multi-Lingual Malaysian Embedding: Leveraging Large Language Models for Semantic Representations},
    pdfauthor={Husein Zolkepli, Aisyah Razak, Kamarul Adha, Ariff Nazhan},
    pdfsubject={Natural Language Processing, Large Language Models, Machine Learning},
    pdfkeywords={language models, malay natural language processing, deep learning},
    colorlinks=true,
    linkcolor=blue,
    citecolor=blue,
    urlcolor=blue
}
\title{Multi-Lingual Malaysian Embedding: Leveraging Large Language Models for Semantic Representations}
\author{
  Husein Zolkepli\thanks{husein@mesolitica.com} \and
  Aisyah Razak\thanks{aisyahrazak171@gmail.com} \and
  Kamarul Adha\thanks{kamarul.adha360@gmail.com} \and
  Ariff Nazhan\thanks{ariffnzhn@gmail.com}
}

\begin{document}

\maketitle

\begin{abstract}
  In this work, we present a comprehensive exploration of finetuning Malaysian language models, specifically Llama2 and Mistral, on embedding tasks involving negative and positive pairs. We release two distinct models tailored for Semantic Similarity and Retrieval-Augmented Generation (RAG).

  For Semantic Similarity, our 600 million parameter Llama2 model outperforms OpenAI text-embedding-ada-002 across all recall@k metrics for b.cari.com.my, c.cari.com.my, Malay news, and Malaysian Twitter test sets.

  In the realm of RAG models, our approach proves competitive with OpenAI text-embedding-ada-002 in the Malaysian context. Notably, our 2 billion parameter Llama2 model achieves superior Recall@5, Recall@10 for the "Melayu" keyword research papers dataset and excels in Recall@3, Recall@5, and Recall@10 for the lom.agc.gov.my dataset.

  These findings underscore the effectiveness of our finetuning strategy and highlight the performance gains in both Semantic Similarity and RAG tasks.

  All models released at \href{https://huggingface.co/collections/mesolitica/malaysian-embedding-6523612bfe5881ad35f81b99}{HuggingFace Mesolitica Malaysian Embedding Collection}.

\end{abstract}

\section{Introduction}

In the wake of the release of ChatGPT, the landscape of conversational AI has seen a surge in the development of chatbots armed with proprietary knowledge bases. The success of these chatbots hinges on their ability to perform semantic search effectively—essentially, retrieving the most relevant information from a knowledge base to provide accurate responses to user queries. This functionality relies on a sophisticated embedding model capable of capturing and understanding the nuances of language.

For English-based applications, companies have readily embraced the closed-source OpenAI text-embedding-ada-002 model as a turnkey solution for robust embedding capabilities. However, when it comes to the Malay language, the performance of such out-of-the-box solutions falls short. The intricacies of Malay linguistics, along with its unique semantic structure, pose challenges that generic models struggle to overcome.

In recognition of this disparity, our research endeavors to address the existing gap in the provision of effective embedding models tailored specifically for the Malay language. Rather than relying on closed-source alternatives, our approach seeks to introduce an open-source solution that not only caters to the linguistic intricacies of Malay but also outperforms existing models in the context of semantic search. By doing so, we aim to propel the field of natural language processing forward, fostering innovation and accessibility for Malay language applications in the realm of conversational AI and knowledge retrieval.

\begin{itemize}
  \item \textbf{Hard mining embedding dataset:} We utilize OpenAI text-embedding-ada-002 and bge-large-en \cite{bge_embedding} from Beijing Academy of Artificial Intelligence as base models to convert Malaysian texts into embedding representations. Subsequently, we employ hard mining techniques to refine and optimize these embeddings, enhancing their quality and relevance. This approach aims to extract more meaningful semantic information from the original texts, ensuring our embeddings align seamlessly with the nuances of the Malay language for improved performance in various applications.

  \item \textbf{Synthetic RAG dataset:} We created positive and negative pairs using synthetic QA dataset from Malaysian Mistral \cite{zolkepli2024large}. These pairs help enhance semantic retrieval, enabling the model to discern intricate connections between contexts and questions. Positive pairs highlight correct retrievals, while negative pairs offer learning opportunities for refining the model's comprehension. This approach enriches the dataset and contributes to developing a more context-aware embedding model for a nuanced understanding of the Malaysian language context.

  \item \textbf{Finetuned Large Language Model:} We fine-tuned Malaysian Llama2 models of 600M, 1B, and 2B parameters using a contrastive loss. To cater to different embedding needs, we extracted the initial N layers of these models and continued pretraining, creating smaller models customized for specific embedding tasks. This approach allows for adaptability and optimal performance in various scenarios.
\end{itemize}

\section{Hard Mining Dataset Procedure}

\subsection{Converting to Embedding representation}

\subsubsection{OpenAI text-embedding-ada-002 base model}

The utilization of OpenAI's text-embedding-ada-002 plays a pivotal role in transforming samples sourced from diverse platforms, including b.cari.com.my, carigold, Malaysian Facebook posts, Lowyat, Malaysian news, and Malaysian Twitter, into rich and meaningful embedding representations. The text-embedding-ada-002 model serves as the cornerstone for capturing the semantic essence of these Malaysian texts, offering a standardized and condensed representation that encapsulates the contextual information present in the original content.

These initial embedding representations serve as the baseline dataset, laying the foundation for our subsequent hard mining process. The diverse array of sources ensures that our baseline dataset is comprehensive, capturing the linguistic nuances and variations prevalent across different Malaysian platforms. By employing text-embedding-ada-002, we aim to establish a robust starting point for the subsequent stages of our embedding model development.

As we embark on the hard mining process, the quality and richness of the baseline dataset become crucial. The embeddings generated by text-embedding-ada-002 provide a solid foundation for identifying and isolating challenging cases within the dataset, setting the stage for the refinement and enhancement of our models through targeted training iterations

All embedding representation dataset and implementation published at \href{https://github.com/mesolitica/malaysian-dataset/tree/master/embedding/ada-002}{mesolitica/malaysian-dataset/embedding/ada-002}.

\subsubsection{bge-large-en base model}

Our approach involves addressing the language diversity challenge posed by bge-large-en, which is specifically trained on English datasets. To overcome this limitation, we implemented a multi-step process to enrich our embedding representation. We initiated the process by generating noisy translations for content from b.cari.com.my, c.cari.com.my, Malaysian Reddit, and Malaysian Twitter. Leveraging the capabilities of ChatGPT3.5, we translated these Malaysian texts into standard English. Subsequently, a post-filtering mechanism was applied to refine and enhance the accuracy of the noisy translations, ensuring that the resulting English translations retained the intended semantic meaning of the original Malaysian content.

The translated segments were then used to create an additional set of embedding representations, adding a layer of linguistic diversity to our dataset. This diversification strategy is particularly important to capture the broad spectrum of linguistic nuances present in Malaysian online platforms. By combining the outputs from bge-large-en and OpenAI text-embedding-ada-002, we aimed to create a more comprehensive and representative embedding model. The synergy of these two models contributes to the holistic coverage of the Malaysian language landscape, enhancing the robustness and inclusivity of our embedding representation dataset. This meticulous process ensures that our models can effectively encapsulate the diversity of language expressions found across various Malaysian online sources.

All embedding representation dataset and implementation published at \href{https://github.com/mesolitica/malaysian-dataset/tree/master/embedding/bge-large-en}{mesolitica/malaysian-dataset/embedding/bge-large-en}.

\subsection {Hard Mining Procedure}

In our analysis, we observed that the distribution of text similarity tends to exhibit a negatively skewed pattern. This indicates that a majority of the generated texts have relatively low similarity scores. Our primary focus lies on the tail-ends of this distribution, particularly on the extreme left-tail, which signifies the most similar texts or what we term as hard positives. On the opposite end, the extreme right-tail represents the most dissimilar texts or hard negatives. By concentrating on these extremes, we aim to capture instances where the texts are either highly similar or distinctly dissimilar, providing a nuanced understanding of the variability within the generated text distribution.

\begin{figure}[h]
  \centering
  \includegraphics[width=0.6\linewidth]{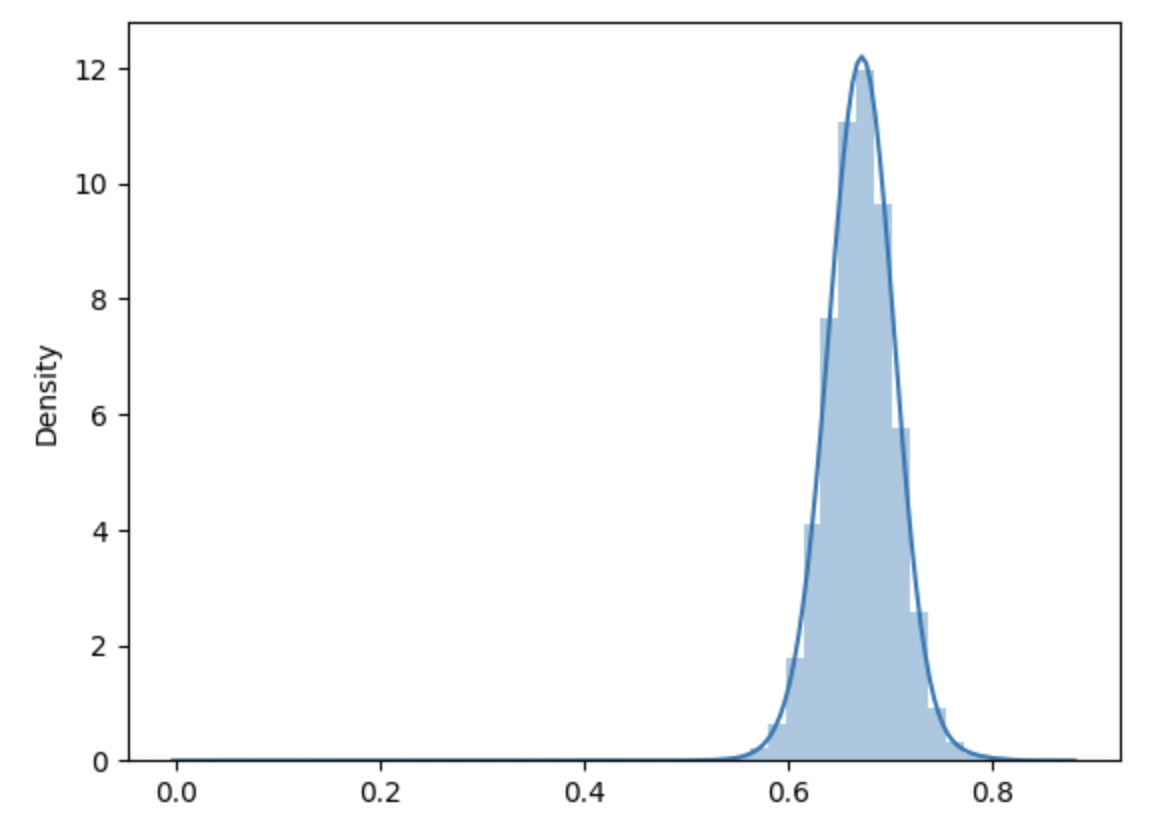}
\end{figure}

The determination of similarity is guided by the Euclidean distance formula, where lower values indicate greater similarity and higher values signify greater dissimilarity. This rigorous process of hard mining allows us to refine and optimize the embeddings, ensuring that they encapsulate the nuanced semantics of the Malay language. This meticulous approach enhances the quality and relevance of the embeddings, resulting in improved performance across a spectrum of applications.

For a given base text like 'Selamat Tidur mutual ku semua,' the 5\% percentile represents the lower end of potential variations. In this instance, the generated results at the 5\% percentile include variations such as 'selamat tidur,' 'selamat tidur katanya,' and 'Selamat tidur!!' These variations capture the diversity in possible renditions, showcasing different expressions and styles within the given context. The percentile-based approach allows for a range of outputs, offering insights into the variability of language generation around the base text.

\pagebreak

In the pursuit of efficient hard mining for our embedding dataset, we strategically employ the scipy.spatial.KDTree approach. While the prevailing trend involves utilizing Faiss multi-GPUs \cite{johnson2017billionscale} for mining tasks, our deliberate choice of scipy.spatial.KDTree stems from a consideration of both effectiveness and cost-effectiveness. Leveraging a virtual machine with ample CPU and memory resources proves to be a more economical alternative compared to configurations with multi-GPUs.

The decision to opt for scipy.spatial.KDTree is rooted in our commitment to resource optimization within the confines of budget constraints. By harnessing the power of this spatial data structure, we strike a balance between achieving effective hard mining and ensuring the judicious utilization of available computing resources. This strategic choice aligns with our overarching goal of delivering optimal performance without compromising on cost efficiency.

Below is pseudo Python code to do hard mining,

\begin{lstlisting}[breaklines=true]
def mining(kd_tree, i, vectors, lower_bound, upper_bound, max_size = 5):
    dist, ind = kd_tree.query(vectors[x], k=len(vectors), workers = 1)

    pos_indices = [k for k in ind[dist <= lower_bound]]
    neg_indices = [k for k in ind[dist > upper_bound]]

    if len(neg_indices) > max_size:
        neg_indices = random.sample(neg_indices,max_size)
    if len(pos_indices) > max_size:
        pos_indices = random.sample(pos_indices,max_size)
\end{lstlisting}

All hard-mining implementation published at \href{https://github.com/mesolitica/llm-embedding/tree/main/mining-openai}{mesolitica/llm-embedding/mining-openai} and \href{https://github.com/mesolitica/llm-embedding/tree/main/mining-bge}{mesolitica/llm-embedding/mining-bge}

\section{Retrieval-Augmented Generation Dataset Procedure}

Addressing the inherent challenges associated with traditional RAG datasets, where the length of context far surpasses that of user questions, necessitates a nuanced strategy to optimize the performance of our language models. In the typical RAG scenario, contexts may span entire documents or articles, providing a rich source of information, while user queries tend to be concise and specific. This incongruence in length poses a considerable hurdle for many existing embedding models, which may struggle to effectively comprehend and respond to extensive contextual information.

In response to this challenge, we have devised a tailored approach leveraging a synthetic Malaysian Open QA dataset from Malaysian Mistral \cite{zolkepli2024large}. This synthetic dataset is curated by supplying contextual information to ChatGPT3.5 and prompting it to generate a comprehensive list of QA pairs based on that context. By employing this method, we ensure that the generated QA pairs align seamlessly with the intricate linguistic nuances of the Malaysian language context, effectively addressing the disparity between the lengths of context and user questions.

Furthermore, we incorporate a robust post-validation mechanism to enhance the overall quality of our dataset. Utilizing a keyword-based validation approach, we meticulously evaluate and accept QA pairs only when the generated answer exhibits a substantial overlap of at least 60\% with the original context. This stringent validation criterion serves as a powerful filter, guaranteeing that the generated QA pairs maintain a high degree of relevance and fidelity to the provided context.

This comprehensive approach, combining synthetic dataset generation and meticulous post-validation measures, empowers our models to effectively navigate and understand extended contextual information. The resulting dataset is finely tuned to the specific requirements of RAG tasks on longer texts, contributing significantly to the enhancement of contextual understanding within our language models.

Example of the dataset,

\begin{lstlisting}[breaklines=true]
  [{'paragraph': 'The Legend of Korra ialah ...',
  'url': 'https://ms.wikipedia.org/wiki?curid=823980',
  'qa': {'qa': [{'question': 'Apakah siri animasi yang ditayangkan di Nickelodeon sejak 2012?',
     'answer': 'The Legend of Korra'},
    {'question': 'Siapakah pencipta siri animasi The Legend of Korra?',
     'answer': 'Bryan Konietzko dan Michael Dante DiMartino'},
    {'question': "Apakah yang dimaksudkan dengan 'bending' dalam siri animasi The Legend of Korra?",
     'answer': 'Kekuatan untuk memanipulasi elemen seperti air, bumi, api, atau udara'},
    {'question': 'Siapakah Avatar Korra?',
     'answer': 'Pengganti Aang dalam siri sebelumnya yang menghadapi pergolakan politik dan roh semangat dalam dunia pemodenan'},
    {'question': 'Apakah kejayaan siri The Legend of Korra?',
     'answer': 'Kejayaan yang kritikal dan komersial dengan jumlah penonton tertinggi bagi siri animasi di Amerika Syarikat pada tahun 2012'}]}},
 {'paragraph': 'adalah sebuah siri televisyen penstriman ...',
  'url': 'https://ms.wikipedia.org/wiki?curid=1070143',
  'qa': {'qa': [{'question': 'Apakah nama siri televisyen tersebut?',
     'answer': 'Alice in Borderland'},
    {'question': 'Siapakah pengarah siri televisyen tersebut?',
     'answer': 'Shinsuke Sato'},
    {'question': 'Apakah tarikh penayangan perdana siri televisyen tersebut di Netflix?',
     'answer': '10 Disember 2020'},
    {'question': 'Apakah ulasan positif yang diterima siri televisyen tersebut?',
     'answer': 'visual, sinematografi, penyuntingan, dan penggunaan grafik kekerasan'},
    {'question': 'Berapa musim siri televisyen tersebut?',
     'answer': 'Dua musim'}]}}]
\end{lstlisting}

The formulation of positive and negative pairs within our dataset intricately contributes to the robustness of our language models. For positive pairs, we draw directly from the actual QA pairs generated based on the provided context. This approach ensures that the positive pairs inherently encapsulate the contextual understanding and thematic relevance of the given information. By anchoring the positive pairs in the real QA generation process, we fortify our dataset with instances that authentically represent the nuanced interplay between context and questions.

The negative pairs play a pivotal role in diversifying the dataset and enhancing the model's ability to discern between relevant and irrelevant information. These negative pairs are sourced from different data points, involving QA pairs generated from alternative contexts. This deliberate introduction of contrasting information encourages the model to refine its discriminatory abilities, discerning not only what constitutes a suitable answer within a given context but also differentiating it from responses generated in dissimilar thematic settings.

By methodically incorporating both positive and negative pairs, our dataset encapsulates a rich spectrum of linguistic intricacies and context-dependent reasoning. This meticulous curation of pairs not only strengthens the contextual understanding of our language models but also fosters a more discerning and adaptive response mechanism, poised to handle a diverse array of user queries within a wide spectrum of contextual scenarios.

Example of generated positive and negative pairs,

\begin{lstlisting}[breaklines=true]
  {'query': 'The Legend of Korra ...',
 'positive_pairs': ['Apakah siri animasi yang ditayangkan di Nickelodeon sejak 2012?',
  'Siapakah pencipta siri animasi The Legend of Korra?',
  "Apakah yang dimaksudkan dengan 'bending' dalam siri animasi The Legend of Korra?"],
 'negative_pairs': ['Apakah tarikh penayangan perdana siri televisyen tersebut di Netflix?',
  'Bilakah pencabaran terhadap Bill Clinton dimulakan?',
  'Pencabaran terhadap Bill Clinton telah dimulakan pada 8 Oktober 1998.']
  }
\end{lstlisting}

All synthetic dataset and implementation published at \href{https://github.com/mesolitica/malaysian-dataset/tree/master/embedding/instructions-pair}{mesolitica/malaysian-dataset/embedding/instructions-pair}.

\section{Fine-tuning Procedure}

\subsection{First N hidden layers Continue Pre-training}\label{sec:n-layers}

We are extending the pretraining of our language models by focusing on the initial N layers from the base models. This helps us create a more compact yet insightful representation. Using the same dataset as Malaysian Mistral \cite{zolkepli2024large}, our approach ensures efficiency. We employ a packing technique with a context length of 32768 to enhance the model's understanding of diverse linguistic contexts.

Our continued pretraining efforts encompass three different parameter sizes: 600 million, 1 billion, and 2 billion. For the 600 million parameter model, we extract information from the first two hidden layers of Malaysian Llama2. Likewise, the 1 billion parameter model draws insights from the first four hidden layers of Malaysian Llama2. Lastly, the 2 billion parameter model leverages knowledge from the first six hidden layers of Malaysian Llama2. This tiered approach allows us to tailor our models to different scales, capturing varying levels of intricacies within the Malaysian language.

During this phase is to maximize the likelihood of the entire sequence,

\[
  P(x_1, x_2, \ldots, x_T) = \prod_{t=1}^{T} P(x_t | x_1, x_2, \ldots, x_{t-1})
\]

We utilized Standard\_NC96ads\_A100\_v4 Azure instance which contains 4x A100 80 GB GPUs. The continue pre-training hyperparameters are detailed below:

\begin{table}[h]
  \centering
  \begin{tabular}{lccl}
    \hline
    \textbf{Hyperparameter} & \textbf{Value} \\
    \hline
    DeepSpeed               & ZeRO-3 Offload \\
    Batch Size              & 4              \\
    Learning Rate           & constant 2e-5  \\
    Precision               & bfloat16       \\
    \hline
  \end{tabular}
\end{table}

Complete continue pre-training 32768 context length implementation at \href{https://github.com/mesolitica/malaya/tree/5.1/session/llama2#600m-32768-context-length-flash-attention-2}{here}.

\subsection{Contrastive Fine-tuning}

In our experimentation, we meticulously curated two distinct datasets to fine-tune our embedding models. The first dataset comprises 100\% hard mining instances, where we leveraged the scipy.spatial.KDTree for efficient mining of relevant pairs. For the second dataset, we embraced a 100\% synthetic approach, generating RAG (Retrieval-Augmented Generation) instances. Notably, 30\% of this synthetic dataset was derived from the hard mining dataset, ensuring a blend of real-world and synthetic data.

For the embedding finetuning process, we opted for a contrastive loss function \cite{1467314}. This choice enables the model to discern between positive and negative pairs effectively. The training sessions were conducted using an 8192 context length, with the flexibility to scale up to 32768 context length seamlessly. This adaptability is crucial as the models were continued pretraining on 32768 context length, contributing to a more robust understanding of longer text inputs.

\begin{equation}
  \mathcal{L}_{\text{contrastive}}(y, d) = \begin{cases}
    (1 - d)^2             & \text{if } y = 1 \\
    \max(d - \alpha, 0)^2 & \text{if } y = 0
  \end{cases}
\end{equation}

For contrastive loss where $y$ is the binary label (1 for positive pairs, 0 for negative pairs), $d$ is the dissimilarity score, and $\alpha$ is a margin parameter.

\pagebreak

The contrastive fine-tuning hyperparameters for 2B parameter model are detailed below:

\begin{table}[h]
  \centering
  \begin{tabular}{lccl}
    \hline
    \textbf{Hyperparameter} & \textbf{Value} \\
    \hline
    DeepSpeed               & ZeRO-3 Offload \\
    Batch Size              & 3              \\
    Learning Rate           & constant 2e-5  \\
    Precision               & bfloat16       \\
    \hline
  \end{tabular}
\end{table}

Complete contrastive fine-tuning 32768 context length implementation at \href{https://github.com/mesolitica/llm-embedding/blob/main/run-2b-contrastive.sh}{here}.

\section{Evaluation}

We meticulously curated test sets from diverse sources, including b.cari.com.my, c.cari.com.my, Malay news, Malaysian Twitter, research papers with the "Melayu" keyword, and the lom.agc.gov.my dataset. These datasets cover a wide spectrum of contexts, reflecting the model's ability to handle various content types and domains.

For the evaluation process, we employed the recall@k metric, which measures the model's effectiveness in retrieving relevant information within the top-k predictions. This evaluation metric allows us to assess how well our models perform in providing accurate and contextually relevant results across different datasets, providing valuable insights into their overall efficacy.

\begin{table}[hbt!]
  \centering
  \begin{tabular}{lcccl}
    \hline
    \textbf{Model}           & \textbf{Recall @1} & \textbf{Recall @3} & \textbf{Recall @5} & \textbf{Recall @10} \\
    \hline
    OpenAI ada-002           & 31.62              & 62.42              & 69.44              & 76.23               \\
    llama2-embedding-600m-8k & 31.68              & 68.81              & 77.89              & 84.53               \\
    llama2-embedding-1b-8k   & \textbf{32.62}     & \textbf{69.47}     & \textbf{79.02}     & \textbf{86.03}      \\
    \hline
  \end{tabular}
  \caption{b.cari.com.my semantic similarity evaluation}
\end{table}

\begin{table}[hbt!]
  \centering
  \begin{tabular}{lcccl}
    \hline
    \textbf{Model}           & \textbf{Recall @1} & \textbf{Recall @3} & \textbf{Recall @5} & \textbf{Recall @10} \\
    \hline
    OpenAI ada-002           & 8.38               & 21.38              & 27.86              & 35.89               \\
    llama2-embedding-600m-8k & 8.32               & 18.73              & 23.97              & 31.4                \\
    llama2-embedding-1b-8k   & \textbf{8.69}      & 19.96              & 26.08              & 34.3                \\
    \hline
  \end{tabular}
  \caption{c.cari.com.my semantic similarity evaluation}
\end{table}

\begin{table}[hbt!]
  \centering
  \begin{tabular}{lcccl}
    \hline
    \textbf{Model}           & \textbf{Recall @1} & \textbf{Recall @3} & \textbf{Recall @5} & \textbf{Recall @10} \\
    \hline
    OpenAI ada-002           & 13.86              & 29.52              & 37.45              & 47.54               \\
    llama2-embedding-600m-8k & 14.26              & 27.58              & 36.4               & 47.11               \\
    llama2-embedding-1b-8k   & \textbf{14.61}     & 28.52              & 36.4               & \textbf{47.75}      \\
    \hline
  \end{tabular}
  \caption{Malay News semantic similarity evaluation}
\end{table}

\begin{table}[hbt!]
  \centering
  \begin{tabular}{lcccl}
    \hline
    \textbf{Model}           & \textbf{Recall @1} & \textbf{Recall @3} & \textbf{Recall @5} & \textbf{Recall @10} \\
    \hline
    OpenAI ada-002           & 22.94              & 49.19              & 59.3               & 72.48               \\
    llama2-embedding-600m-8k & \textbf{23.49}     & 52.0               & 64.16              & 78.59               \\
    llama2-embedding-1b-8k   & 23.05              & \textbf{53.42}     & \textbf{64.98}     & \textbf{79.25}      \\
    \hline
  \end{tabular}
  \caption{Malaysian Twitter semantic similarity evaluation}
\end{table}

\begin{table}[hbt!]
  \centering
  \begin{tabular}{lccccl}
    \hline
    \textbf{Model}           & \textbf{Recall @1} & \textbf{Recall @3} & \textbf{Recall @5} & \textbf{Recall @10} \\
    \hline
    OpenAI ada-002           & 31.6               & 51.2               & 58.78              & 67.21               \\
    llama2-embedding-600m-8k & 12.6               & 26.5               & 35.35              & 47.51               \\
    llama2-embedding-1b-8k   & 22.0               & 42.8               & 53.79              & 67.08               \\
    llama2-embedding-2b-8k   & 26.7               & 49.1               & \textbf{60.91}     & \textbf{73.97}      \\
    \hline
  \end{tabular}
  \caption{"Melayu" keyword research papers RAG evaluation}
\end{table}

\begin{table}[hbt!]
  \centering
  \begin{tabular}{lccccl}
    \hline
    \textbf{Model}           & \textbf{Recall @1} & \textbf{Recall @3} & \textbf{Recall @5} & \textbf{Recall @10} \\
    \hline
    OpenAI ada-002           & 19.16              & 28.27              & 32.25              & 36.85               \\
    llama2-embedding-600m-8k & 6.6                & 13.75              & 18.86              & 27.45               \\
    llama2-embedding-1b-8k   & 11.89              & 23.97              & 30.94              & 41.34               \\
    llama2-embedding-2b-8k   & 15.49              & \textbf{29.11}     & \textbf{37.25}     & \textbf{48.86}      \\
    \hline
  \end{tabular}
  \caption{lom.agc.gov.my RAG evaluation}
\end{table}

\pagebreak

We also compared with other models and published the benchmark at \href{https://huggingface.co/spaces/mesolitica/malaysian-embedding-leaderboard}{mesolitica/malaysian-embedding-leaderboard}.

\section{Acknowledgement}

Special thanks to Malaysia-AI volunteers especially \href{https://www.linkedin.com/in/wan-adzhar-faiq-adzlan-19a27baa/}{Wan Adzhar Faiq Adzlan}, \href{https://www.linkedin.com/in/ammar-azman/}{Ammar Azman}, \href{https://www.linkedin.com/in/amzar96/}{M. Amzar}, \href{https://www.linkedin.com/in/muhammad-farhan-helmy-0529501a7/}{Muhammad Farhan} and \href{https://www.linkedin.com/in/syafie-nizam/}{Syafie Nizam} for contributing dataset to train Malaysian Embedding models.

We would like to express our gratitude to NVIDIA Inception for generously providing us with the opportunity to train our model on the Azure cloud. Their support has played a crucial role in the success of our research, enabling us to leverage advanced technologies and computational resources.

We extend our thanks to the wider research community for their valuable insights and collaborative discussions, which have greatly influenced our work. This paper reflects the collective efforts and contributions from both NVIDIA Inception and the broader research community.

\section{Conclusion}

In conclusion, we have introduced an open-source Malaysian embedding model that exhibits competitive performance, particularly excelling in tasks such as Retrieval-Augmented Generation (RAG) and semantic similarity. This model stands as a viable alternative, eliminating the reliance on closed-source solutions. By offering a publicly accessible and proficient Malaysian embedding model, our contribution aims to foster transparency, accessibility, and innovation within the field, paving the way for diverse applications and further advancements in natural language processing for the Multi-language in Malaysia.

\bibliography{neurips_2023}{}

\begin{thebibliography}{1}

\bibitem{bge_embedding}
Shitao Xiao, Zheng Liu, Peitian Zhang, and Niklas Muennighoff.
\newblock C-pack: Packaged resources to advance general chinese embedding,
  2023.

\bibitem{zolkepli2024large}
Husein Zolkepli, Aisyah Razak, Kamarul Adha, and Ariff Nazhan.
\newblock Large malaysian language model based on mistral for enhanced local
  language understanding, 2024.

\bibitem{johnson2017billionscale}
Jeff Johnson, Matthijs Douze, and Hervé Jégou.
\newblock Billion-scale similarity search with gpus, 2017.

\bibitem{1467314}
S.~Chopra, R.~Hadsell, and Y.~LeCun.
\newblock Learning a similarity metric discriminatively, with application to
  face verification.
\newblock In {\em 2005 IEEE Computer Society Conference on Computer Vision and
  Pattern Recognition (CVPR'05)}, volume~1, pages 539--546 vol. 1, 2005.

\end{thebibliography}
\bibliographystyle{unsrt}

\end{document}